\pdfoutput=1

\documentclass[11pt]{article}

\usepackage[]{acl}

\usepackage{times}
\usepackage{latexsym}

\usepackage[T1]{fontenc}

\usepackage[utf8]{inputenc}

\usepackage{microtype}

\usepackage{inconsolata}
\usepackage{graphicx}
\usepackage{bm}
\usepackage{amsmath}
\usepackage{amsfonts}
\usepackage{xspace}
\usepackage{multicol}
\usepackage{multirow}
\usepackage{subfigure}
\usepackage{makecell}
\usepackage{listings}
\usepackage{adjustbox}
\usepackage{booktabs}
\usepackage{hyperref}
\usepackage[ruled, vlined, linesnumbered]{algorithm2e}

\definecolor{codegreen}{rgb}{0,0.6,0}
\definecolor{codegray}{rgb}{0.5,0.5,0.5}
\definecolor{codepurple}{rgb}{0.58,0,0.82}
\definecolor{backcolour}{rgb}{0.95,0.95,0.92}

\lstdefinestyle{mystyle}{
    backgroundcolor=\color{backcolour},   
    commentstyle=\color{codegreen},
    stringstyle=\color{codepurple},
    basicstyle=\ttfamily\scriptsize,
    breakatwhitespace=true,         
    breaklines=true,                 
    captionpos=b,                    
    keepspaces=true,                 
    numbers=none,                    
    numbersep=5pt,                  
    showspaces=false,                
    showstringspaces=false,
    showtabs=false,                  
    tabsize=2,
    columns=flexible,
    escapeinside={(*}{*)},
}

\usepackage{enumitem}
\usepackage{caption}

\newcommand{\nonfirstparagraph}[1]{\paragraph{#1}}

\newcommand{\our}{X-MLClass\xspace}

\title{Open-world Multi-label Text Classification with\\ Extremely Weak Supervision}

\author{
Xintong Li$^1$, Jinya Jiang$^1$, Ria Dharmani$^1$ \\ \bf Jayanth Srinivasa$^2$, Gaowen Liu$^2$, Jingbo Shang$^1$ \\
University of California, San Diego$^1$ \quad Cisco$^2$ \\
\texttt{\{xil240, j9jiang, rdharmani, jshang\}@ucsd.edu}\\
\texttt{\{jasriniv, gaoliu\}@cisco.com}
}

\begin{document}
\maketitle
\begin{abstract}
We study open-world multi-label text classification under extremely weak supervision (XWS), where the user only provides a brief description for classification objectives without any labels or ground-truth label space. 
Similar single-label XWS settings have been explored recently, however, these methods cannot be easily adapted for multi-label. 
We observe that (1)~most documents have a dominant class covering the majority of content and (2)~long-tail labels would appear in some documents as a dominant class.
Therefore, we first utilize the user description to prompt a large language model (LLM) for dominant keyphrases of a subset of raw documents, and then construct a (initial) label space via clustering. 
We further apply a zero-shot multi-label classifier to locate the documents with small top predicted scores, so we can revisit their dominant keyphrases for more long-tail labels.
We iterate this process to discover a comprehensive label space and construct a multi-label classifier as a novel method, \our.
\our exhibits a remarkable increase in ground-truth label space coverage on various datasets, for example, a 40\% improvement on the AAPD dataset over topic modeling and keyword extraction methods.
Moreover, \our achieves the best end-to-end multi-label classification accuracy.

\end{abstract}

\begin{figure*}[t]
    \centering
    \includegraphics[width = 0.98 \linewidth]{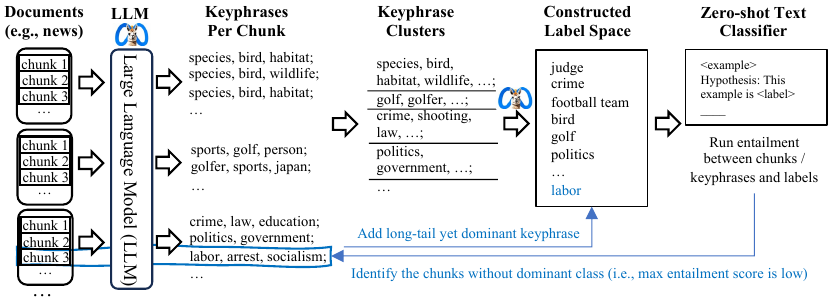}
     \caption{An overview of our \our framework. 
     The only required supervision from the user is a brief description of the classification objective.
     During the first LLM prompting stage for keyphrases, \our leverages this description as a part of the prompt, so it will be helpful if the description includes some demonstrations.
     }
    \label{fig:overview}
\end{figure*}
\section{Introduction}
\label{sec:intro}

Multi-label text classification (MLTC) aims to assign one or more labels to each input document in the corpus. Traditional methods~\cite{liu2022bertflowvae, Xiong2021ExtremeZL, gera2022zero} require a complete list of class names, which is challenging to provide beforehand given the massive number of documents and diverse topics. 
This work focuses on a new problem, open-world\footnote{Our ``open-world” definition denotes the absence of any class information during the training and testing phase, which is more challenging than the traditional settings.} MLTC
under extremely weak supervision (XWS), 
where the user only provides a brief description for classification objectives without any labels or ground-truth label space. 
Despite the considerable technical challenges posed by this task, its practical significance in real-world applications cannot be underestimated. 
For instance, the need for product tagging and categorization is ubiquitous in online shopping platforms. 
It requires identifying multiple labels for each product without access to a predefined label space, a challenge our model adeptly addresses.

The most related XWS problems are text clustering~\cite{Zhang2023ClusterLLMLL, Wang2023GoalDrivenEC} and topic modeling~\cite{Grootendorst2022BERTopicNT, Pham2023TopicGPTAP}, where those methods are typically only capable of assigning a single label to each document.
These single-label methods cannot be easily adapted for multi-label. 

We observe that (1)~most documents have a dominant class covering the majority of content and (2) long-tail labels would appear as the dominant class in some documents. Experiments reveal that over 90\% of documents contain a dominant class, and 100\% of labels appear as the dominant class in at least one document, as analyzed further in Section~\ref{sec:observation}.
Based on these observations, we propose a novel method, \our, to discover a pragmatic label space by iteratively adding (long-tailed) labels and construct a multi-label text classification classifier with the assistance of LLM (i.e., \texttt{llama-2-13b-chat} in our experiments), as shown in Figure~\ref{fig:overview}. 
Our approach requires only a brief user description about the classification objective as prompt for LLM, which significantly reduces the cost of model training. 

The first step in \our is to construct a high-quality label space.
We start with a reasonably large subset of the raw documents.
For each document, we partition it into chunks to better align with the context length of LLM and also ensure that each chunk contains a single topic, and then prompt the LLM to generate the most dominant keyphrases for each chunk.
As previous LLM-based text clustering work has suggested~\cite{Wang2023GoalDrivenEC,wang2023wot}, there are very likely some semantically redundant yet lexically different keyphrases among the generated ones. 
We cluster these keyphrases, and within every cluster, we pull together the corresponding chunks of the keyphrases closest to the cluster center to prompt the LLM once again, generating one single label for each cluster. 
After eliminating labels with high similarity scores, we constitute an initial label space.

We then apply the state-of-the-art textual entailment-based classification methods~\cite{pamies2023weakly} to construct a classifier  that revisits the documents and identifies long-tail labels. 
Specifically, we query every text chunk against all the labels for the entailment score, identifying chunks with low top predicted scores that lack a dominant class.
We revisit the keyphrases generated by these chunks to unveil more long-tail labels, selectively choosing keyphrases with a modest presence in the entire keyphrase set but absent in the original label space.
These new keyphrases are included in the label set, and documents are reassessed with this updated set. By repeating these steps for a fixed number of iterations, the final label space contains a substantial number of long-tail labels.

Extensive experiments on 5 benchmark datasets reveal the superiority of \our outperforming all compared methods. 
Remarkably, compared with baselines, \our achieves a significant enhancement of 40\% and 30\% in ground-truth label space coverage on the AAPD and RCV1-V2 datasets. 
Furthermore, it achieves the highest accuracy in zero-shot MLTC, surpassing the top-ranking models on HuggingFace across all datasets.  

Our contributions are summarized as: 
\begin{itemize}[nosep,leftmargin=*]
    \item We attack a challenging problem, open-world MLTC with XWS, 
    where the user only provides a brief description for classification objectives without any labels or ground-truth label space. 
    \item 
    We propose a novel framework \our which iteratively discovers the label space and builds an MLTC classifier.
    \item Compared with all traditional label generation methods, \our achieves a significantly higher coverage score along with superior end-to-end classification accuracy.
\end{itemize}
Our source code and dataset can be obtained here~\footnote{\url{https://github.com/Kaylee0501/X-MLClass}}.

\section{Related Work}

\paragraph{Topic Modeling:} Topic modeling has been widely adopted for discovering latent thematic structures within collections of text documents. Traditional models, such as Latent Dirichlet Allocation (LDA)~\cite{blei2003latent} and Non-Negative Matrix Factorization (NMF)~\cite{fevotte2011algorithms} represent documents as mixtures of latent topics using bag-of-words representations. New techniques like Top2Vec~\cite{angelov2020top2vec} and BERTopic~\cite{Grootendorst2022BERTopicNT} build primarily on clustering embeddings, demonstrating the potential of embedding-based topic modeling approaches. Another recent method, TopicGPT~\cite{Pham2023TopicGPTAP}, takes a different approach by prompting large language models for topic generation, aligning more closely with ground truth labels. However, these existing methods typically provide a single topic for each document, which poses challenges when extending them to multi-label scenarios.

\nonfirstparagraph{Multi-label Text Classification:}  Numerous approaches have been proposed to tackle the complexities of Multi-Label Text Classification (MLTC) problems. Bhatia and Jain~\cite{bhatia2015locally} employ embedding-based methods, leveraging the power of embeddings to train individual classifiers for each label. Later, XML-CNN~\cite{liu2017deep} uses a Convolutional Neural Network (CNN) to learn text representations, demonstrating improvements in MLTC accuracy. Recent works have started to tackle MLTC problems using a small amount of labeled data or even with no labels at all. For example, \citet{shen-etal-2021-taxoclass} achieves impressive results by using only class names and taxonomies. \citet{rios-kavuluru-2018-shot} train a neural architecture with both true labels and their natural language descriptor. However, these methods still require access to the ground-truth label space.

\nonfirstparagraph{Open-world Single-label Text Classification:}
There has been a surge in open-world models utilizing LLM prompts to derive labels without relying on ground-truth label spaces. Notably, \textsc{GOALEX}~\cite{Wang2023GoalDrivenEC} generates labels for text samples based on users' specific goals, demonstrating a goal-driven approach.
Another noteworthy model, \textsc{ClusterLLM}~\cite{Zhang2023ClusterLLMLL}, leverages API-based LLMs to guide text clustering, resulting in improved performance.
The approach of intent discovery~\cite{zhang2022new}, aiming to infer latent intents from a document set, has proven effective in generating label spaces. A newly introduced method, \textsc{IDAS}~\cite{de2023idas}, prompts LLMs to succinctly summarize utterances, enhancing intent prediction.

\section{Problem Formulation}\label{sec:foundation}

Given an unlabeled corpus $\mathcal{D}$ $=$ $\{D_1$, $D_2$, $\dots$, $D_n\}$, where $D_i \in \mathcal{D}$ represents a document in the collection. 
Our task is to (1) identify class names $\mathcal{C} = \{C_j\}_{j =1}^K$, where $K$ is the \emph{unknown} number of classes, and (2) build a text classifier $f(\cdot)$ to map any raw document $D_i$ to its target labels $Y_i = \{y_i^j\}_{j =1}^p$, where $y_i^j$ is the single label name and $p$ is the number of target labels for $D_i$.
The definition of "open world" denotes the absence of labeled information during training, which is more stringent than the original definition where new labels only appear in the test phase. Several existing works~\cite{brunet2023icl, Zhu2023TextAI} adopt a similar ``open world'' definition to ours.

To the best of our knowledge, this is the first work that explores open-world multi-label text classification without the presence of a ground-truth label space. 
Given the challenging nature of the problem, we assume that human experts are willing to devote some very limited effort, i.e., \emph{extremely weak supervision}, typically in the form of brief classification objective descriptions.

\section{Our Observations}\label{sec:observation}

Our two observations mentioned in Section~\ref{sec:intro} are confirmed by experiments based on 5 benchmark datasets: AAPD~\cite{Yang2018SGMSG}, Reuters-21578~\cite{Sebastiani2003AnAO}, RCV1-V2~\cite{lewis2004rcv1}, DBPedia-298~\cite{lehmann2015dbpedia}, and Amazon-531~\cite{McAuley2013HiddenFA}. 
Specifically, we prompt a large language model (LLM) to check if any of the ground truth labels of a given document is dominant, i.e., covering more than 50\% of the content; and if it exists, which one is the dominant label.~\footnote{The specific prompt can be found in Appendix~\ref{sec:prompt}}

We randomly sample 1000 examples from each dataset and calculate the percentage of documents with a dominant class.
The dominance percentages across datasets are AAPD: 92\%, RCV1-V2: 91.7\%, DBPedia-298: 95.2\%, Reuters-21578: 96.4\%, and Amazon-531: 87.3\%. 
Notably, Reuters dataset exhibits a higher proportion of dominant labels due to its mainly consisting of single-labeled examples. Conversely, Amazon presents a lower dominance percentage, attributed to the mix of fine-grained and coarse-grained labels, posing challenges in determining dominance.
Overall, our analysis indicates that across all datasets, more than 90\% of documents contain a dominant class.
Moving to our second observation, labels existing in less than 1\% of the dataset are identified as long-tail labels. Upon examination, instances of long-tail labels emerging as dominant classes in at least one document are observed, indicating that 100\% of labels serve as dominant classes in some instances.
These observations highlight the potential for generating labels, particularly long-tail labels, from raw documents and guide the design of our framework.

\section{Our \our Framework}

\our consists of three key steps. 
First, every document is split into chunks and transformed into keyphrases by prompting an LLM to construct an initial label space. We further assign labels to each raw document $D_i$ using a custom keyphrase-chunk zero-shot textual entailment classifier. Finally, we iteratively enhance the label space by incorporating additional long-tail labels.  
The framework overview is depicted in Figure~\ref{fig:overview}, and the below sections provide a detailed discussion. 

\subsection{Initial Label Space Construction}

The first step in \our is to construct a high-quality label space.
To balance label coverage and the computational cost of LLM, \our is applied to a reasonably large subset of the corpus $\mathcal{D}$, denoted as $\mathcal{D}_{sub} \subset \mathcal{D}$.

\paragraph{Dominant Keyphrase Generation:} 
For each document, we partition it into chunks to better align with the context length of LLM, and then prompt for the most dominant keyphrases per chunk.
Specifically, each document $D_i \in \mathcal{D}_{sub}$ is segmented into chunks $ \{S_i^1, S_i^2, \dots\}$, with a predefined chunk size of 50 tokens. 
This choice is also made to ensure each chunk primarily contains one label.
To generate keyphrases for each chunk $S_i^j$, we employ an LLM and provide it with an instruction based on a brief user description of the classification objective.~\footnote{The specific prompt can be found in Appendix~\ref{sec:phrases}}
The LLM then refines keyphrases $p_i^j$ from the chunk $S_i^j$, serving as potential class candidates for subsequent stages of our \our model. 
Keyphrases generated from each chunk collectively form a set $\mathcal{P}$.

\paragraph{Keyphrase Clustering:}
As previous LLM-based text clustering work has suggested~\cite{Wang2023GoalDrivenEC,wang2023wot}, there are very likely some semantically redundant yet lexically different keyphrases among the generated ones, so 
we employ the instruction-tuned text embedding model, \texttt{instructor-large}~\cite{su2022one}, to generate vector representations for all the keyphrases in $\mathcal{P}$. 
Traditional clustering methods face challenges in high-dimensional spaces~\cite{aggarwal2001surprising, wang2020x}.  To address this, we apply UMAP~\cite{mcinnes2018umap} for dimensionality reduction, effectively balancing local and global structures.  
Finally, we obtain the clusters using the Gaussian Mixture Model (GMM) in the projected low-dimensional space, renowned for its enhanced flexibility in capturing intricate data distributions.

\paragraph{Number of Clusters:}
The number of clusters is determined by considering both the insights of human experts regarding the magnitude of the label space and non-parametric clustering methods such as BERTopic~\cite{Grootendorst2022BERTopicNT}, a highly effective topic modeling method. 
For example, one can train BERTopic on the keyphrase set $\mathcal{P}$ to obtain the topic number $K^0$, serving as the hyper-parameter to GMM. 
This approach also ensures a fair comparison with baseline methods by maintaining consistency in the number of clusters.

\paragraph{Redundant Keyphrase Removal:}
Within every cluster, we focus on the three keyphrases closest to the cluster center to synthesize one single label.
Instead of directly employing the keyphrases for label space creation, we trace back to the original chunks that generated these keyphrases, as they likely contain similar content and represent the same label.
Concatenating these three chunks for each cluster results in a new document. 
For each document, we prompt LLM with an instruction  ``\textit{find one label for this document}'', yielding the initial $K^0$ classes $\{C_j\}_{j =1}^{K^0}$. 
This strategy allows us to generate a single label that best represents the cluster.

This initial label space may contain redundant labels. Sentence-Transformer models~\cite{wang2020minilm} are used to identify distinct pairs of classes with relatively high cosine similarity scores. 
The first class in each identified pair is then removed to eliminate redundancy.
For those borderline similar label pairs, we prompt \texttt{GPT-4}~\cite{achiam2023gpt} with an instruction ``\textit{Do label pairs have similar meanings? If Yes, please output the label that we should delete}.'' to help us detect distinct pairs where cosine similarity scores alone are insufficient. 
This method proves effective in creating a robust label space $\{C_j\}_{j =1}^{K^1}$, and while human involvement can enhance the refinement process, it is not mandatory. 
Further details on human involvement in the label space refinement are provided in appendix~\ref{sec:human-involve}.

\subsection{Textual Entailment-based Classifier}

Given a label space, we build a zero-shot textual entailment-based classifier~\cite{yin2019benchmarking}. 
Since each chunk is designed to have only one label, state-of-the-art zero-shot single-label text classification methods~\cite{pamies2023weakly, gera2022zero, he2021debertav3} are all applicable here.
Specifically, we compare every text chunk against all the labels using a textual entailment model.
For each chunk $s \in \mathcal{S}$ and each class name $c \in \mathcal{C}$, we derive $E_{s,c}$ representing the confidence for the chunk $s$ entailing the hypothesis ``\textit{This example is constructed for $c$}'', and similarly obtain $E_{p,c}$ for each keyphrase $p \in \mathcal{P}$.
Subsequently, for each example in $\mathcal{S}$, we identify the label $c^*$ with the top entailment score, denoted by $E_{s,c^*} > E_{s,c}, \forall c \neq c^*$. 

Finally, we find all $s$ and $p$ belong to the same document $D_i$ and group them into a new set $Q$. For each instance in $Q$, we rank the label candidates according to their entailment scores.   We identify the labels that occur most frequently with the same ranking as the predicted labels for document $D_i$, progressing from the top-ranking to the lowest-ranking order.

\subsection{Label Space Improvement}
We further identify the chunks with lower top predicted scores --- these chunks lack a dominant class in the initial label space. By ranking $E_{s, c^*}$ in ascending order, we select a subset $\mathcal{S}_{sub} \subset \mathcal{S}$ with relatively lower entailment scores, suggesting potential association with tail classes not included in our label space. 
Since keyphrases generated by the LLM may include outliers that are too specific to their corresponding documents, we retain only keyphrases occurring more than 15 times.
For each $s \in \mathcal{S}_{sub}$, we examine all keyphrases in the corresponding $p$. If a keyphrase is absent in the label space but occurs more than 15 times in $\mathcal{P}$, we incorporate it into the label space $\mathcal{C}$.  

Additionally, we compute the frequency of each label $c$ with the top entailment score. 
Labels with lower frequency are removed from the label space $\mathcal{C}$.
The high-frequency labels, secured as a part of the label space, are temporarily excluded from the later label space improvement process. By iteratively training the classifier based on the updated label space, the label set gets finalized by adding more long-tail labels. In the concluding stages, all high-frequency labels are reintroduced, culminating in the formation of ultimate label space.

\begin{table}[t]
\small
  \begin{center}
    \begin{tabular}{lrrr} 
    \toprule
      \textbf{Dataset} & \textbf{\#\,Train} & \textbf{\#\,Text}  & \textbf{\#\,Class}\\
      \midrule
      AAPD & 53,840 & 2,000 & 54\\
      Reuters & 7,769 & 3,019 & 90\\
      RCV1-V2 & 643,531 & 160,883 & 103\\
      DBPedia & 196,665 & 49,167 & 298\\
      Amazon & 29,487 & 19,658 & 531\\
      \bottomrule
    \end{tabular}
    \vspace{-3mm}
    \caption{Dataset statistics.}
    \label{table:datasets}
    \vspace{-8mm}
  \end{center}
\end{table}

\section{Experiments}

We assess the performance of \our through two primary criteria: label space quality and zero-shot MLTC accuracy. 
Our evaluation involves a comparison of our model's label coverage with that of four topic modeling and three keyword extraction methods. 
In terms of end-to-end classification accuracy, we test our method with several top-ranking models available on HuggingFace. 
The subsequent section provides comprehensive details on the datasets, baseline methods, evaluation metrics, implementation specifics, and performance analysis.

\subsection{Datasets}
We perform experiments on five benchmark datasets for multi-label text classification across various domains: \textbf{AAPD}, \textbf{Reuters}-21578, \textbf{RCV1-V2}, \textbf{DBPedia}-298, and \textbf{Amazon}-531. Detailed information about each dataset is provided in Appendix~\ref{sec:dataset}.
Table~\ref{table:datasets} shows that the number of labels in these datasets varies from tens to hundreds.
All the methods will be applied on the documents from the training set, and then evaluated on the test set.

\subsection{Compared Methods}

We compare our \our framework with two types of methods.

\paragraph{Label (Space) Generator:}
We select four representative \emph{topic modeling} methods with distinct paradigms. 
\textbf{LDA}~\cite{blei2003latent} and \textbf{NMF}~\cite{fevotte2011algorithms} extract topics based on word frequency within documents.  
\textbf{Topic2Vec}~\cite{angelov2020top2vec} extends the Word2Vec model to embed topics, facilitating the exploration of semantic relationships between documents. 
Meanwhile, \textbf{BERTopic}~\cite{Grootendorst2022BERTopicNT} leverages BERT embeddings and the HDBSCAN clustering algorithm to identify topics.

The three \emph{keyword extraction} methods include \textbf{PKE}\cite{boudin2016pke}, \textbf{TextRank}\cite{mihalcea2004textrank}, and \textbf{TF-IDF}~\cite{ramos2003using}. PKE selects keyphrase candidates based on their confidence scores, TextRank is a graph-based ranking model that ranks keywords using a voting mechanism, and TF-IDF scores words based on their frequency in the document and inversely proportional to their frequency across documents.

Despite the above methods generating a single label per document, based on our observations, they often align with the dominant label for each document. Given their potential to cover all labels, it is reasonable to compare the label space coverage between our task and these label generators.

\paragraph{Zero-shot Text Classification:}
State-of-the-art zero-shot text classifiers typically follow textual entailment~\cite{yin2019benchmarking, pamies2023weakly}.
Therefore, we choose three entailment models: 
(1) \textbf{bart-large-mnli} exclusively trained on the MNLI dataset,
(2) \textbf{deberta-v3-large-all} trained on 33 datasets reformatted into the universal NLI format,
and (3) \textbf{xlm-roberta-large-xnli} fine-tuned on the XNLI dataset.  
We apply these models using the HuggingFace Transformer pipeline, with a hypothesis template ``\textit{This example is \{label\}}''.

\subsection{Evaluation Metrics}

\paragraph{Label Space Quality:}
We employ an automatic evaluation metric, \textbf{coverage}, to meature the alignment between the ground-truth (GT) label space and the predicted label space. 
A ground-truth label is considered ``covered'' if it exhibits a similarity score exceeding a predefined threshold when compared to a predicted label, or if it receives a positive evaluation from GPT-4. 
We compute the similarity scores using the \texttt{all-MiniLM-L6-v2} model from HuggingFace Sentence-Transformers.
If the similarity score exceeds 0.75, the ground-truth label is considered ``covered.'' For scores between 0.5 and 0.75, we prompt GPT-4 with specific instructions (see Appendix~\ref{sec:gpt}). Labels with scores below 0.5 are not compared due to obvious dissimilarity.

The coverage score is computed as follows:
\begin{equation*}
\text{Coverage} = \frac{1}{N} \mathbb{G}\left(\mathbb{I}\left({C}^{\text{pred}},C^{\text{GT}}\right)\right)
\end{equation*} 
Here $N$ is the total number of topics in the GT label set, $C^\text{GT} (C^\text{pred})$ denotes the set of \text{ground-truth}~(\text{predicted}) labels. 
$\mathbb{I}$ is an indicator that returns 1 if the GT label is considered ``covered''. 
$\mathbb{G}$ represents the bipartite graph maximum match algorithm.

\begin{table}[t]
\small
  \begin{center}
  \setlength{\tabcolsep}{0.9mm}{
    \vspace{-3mm}
\resizebox{\linewidth}{!}{
    \begin{tabular}{l rrrrr} 
    \toprule
      \textbf{Model} & \textbf{AAPD} & \textbf{Reuters} & \textbf{RCV1-V2} & \textbf{DBPedia} & \textbf{Amazon}\\
      \midrule
      LDA       & 30.61 & 17.77 & 12.74 & 11.40 & 13.18\\
      NMF       & 28.57 & 17.77 & 12.74 & 28.18 & 15.06\\
      Top2Vec   & 32.65 & 20.00 & 28.43 & 30.87 & 14.50\\
      BERTopic  & 20.41 & 20.00 & 10.68 & 33.22 & 16.76\\
      PKE  & 14.29 & 14.44 & 25.24 & 25.84 & 15.44\\
      TextRank  & 18.37 & 11.11 & 11.65 & 14.77 & 12.43\\
      TF-IDF  & 14.29 & 15.56 & 8.74 & 26.51 & 12.24\\
      \our & \textbf{77.56} & \textbf{37.78} & \textbf{61.17} & \textbf{67.45} & \textbf{38.04}\\
      \bottomrule
  \end{tabular}
}
\caption{Label Space Coverage Comparison. 
    Top2Vec and BERTopic generate topics with multiple keywords. 
    The predicted label is determined by selecting the top-ranking keyword based on each model's setting.}
\label{table:coverage}
}
  \vspace{-5mm}
  \end{center}
\end{table}

\paragraph{Classification Accuracy:}
Because of the large label space, multi-label text classification typically employs the rank-based evaluation metric precision at $k$, i.e., $\textrm{P@k}$.
It captures the percentage of true labels among top-$k$ score labels and is used for performance comparison. 
$\textrm{P@k}$ can be defined as:
\begin{equation*}\label{equation_silhouette}
\textrm{P@k} = \frac{1}{N} \sum_{i= 1}^N \frac{C_i^{r_k} \cap L_i}{\min(k, |L_i|)}
\end{equation*}
where $L_i$ and $C_i$ denote the true labels and predicted labels for document $D_i$, $|L_i|$ is the number of true labels for $D_i$, and $r_k$ is the $k$-th highest predicted label. 
The ground-truth label is defined ``covered'' using the same \textbf{coverage} metric.

\begin{table*}[t]
\centering
\small
\vspace{-3mm}
\setlength{\tabcolsep}{1.5mm}{
\begin{tabular}{l@{\hspace{0.1cm}}l cc cc cc cc cc}
\toprule
\multirow{2.5}[0]{*}{\textbf{Classifier}} & \multirow{2.5}[0]{*}{\textbf{Method}} & \multicolumn{2}{c}{\textbf{AAPD}}  & \multicolumn{2}{c}{\textbf{Reuter}} & \multicolumn{2}{c}{\textbf{RCV1-V2}} & \multicolumn{2}{c}{\textbf{DBPedia}} & \multicolumn{2}{c}{\textbf{Amazon}}\\ 
\cmidrule(lr){3-4} \cmidrule(lr){5-6} \cmidrule(lr){7-8} \cmidrule(lr){9-10} \cmidrule(lr){11-12}
 & & P@1 & P@3  & P@1 & P@3  & P@1 & P@3  & P@1 & P@3  & P@1 & P@3\\
\midrule
\multirow{2}[0]{*}{\makecell[l]{\texttt{bart-large-}\\\texttt{mnli}}} & Vanilla  & 0.2030 & 0.2585 & 0.0940 & 0.2547 & 0.4220 & 0.3760 & 0.6420 & 0.3657  & 0.5080 & 0.3930 \\ 
\cmidrule{2-12}
 & \our & \textbf{0.3323} & \textbf{0.3735} & \textbf{0.1630} & \textbf{0.3617} & \textbf{0.4530} & \textbf{0.3808} & \textbf{0.6890} & \textbf{0.4000} & \textbf{0.6620} & \textbf{0.4608} \\ 
\midrule
\multirow{2}[0]{*}{\makecell[l]{\texttt{deberta-v3-}\\\texttt{large-all}}}& Vanilla & \textbf{0.3860} & 0.3700 & 0.1290 & 0.3937 & 0.4660 & 0.4152 & \textbf{0.6370} & 0.3816  & 0.5970 & 0.4068  \\ 
\cmidrule{2-12}
 & \our & 0.3573 & \textbf{0.4009} & \textbf{0.1320} & \textbf{0.4150} & \textbf{0.4820} & \textbf{0.4307} & 0.6350 & \textbf{0.3953} & \textbf{0.6200} & \textbf{0.4650} \\ 
\midrule
\multirow{2}[0]{*}{\makecell[l]{\texttt{xlm-roberta-}\\\texttt{large-xnli}}}& Vanilla & 0.1860 & 0.2330 & 0.1450 & 0.3477 & 0.3270 & 0.3053 & 0.6500 & 0.3673  & 0.5060 & 0.3860  \\ 
\cmidrule{2-12}
 & \our & \textbf{0.2713} & \textbf{0.3467} & \textbf{0.2040} & \textbf{0.3968} & \textbf{0.4040} & \textbf{0.3383} & \textbf{0.6610} & \textbf{0.3910} & \textbf{0.5840}  & \textbf{0.4547}\\ 
\bottomrule
\end{tabular}
}
\vspace{-3mm}
\caption{Zero-Shot Multi-Label Text Classification Accuracy Comparison: Vanilla method represents baseline classifier trained on raw documents. We compare the Vanilla performance with \our performance. 
}
\label{table:accuracy}
\vspace{-3mm}
\end{table*}

\subsection{Implementation Details}   
We use \texttt{llama-2-13b-chat} LLM for \our implementation.
The chunk size is uniformly set to 50 across all datasets, ensuring a consistent approach. 
In configuring LDA and NMF, we align the number of topics with our approach. For Top2Vec and BERTopic, which employ HDBSCAN as a clustering method, specifying an exact number of topics is not feasible. However, to maintain consistency, we ensure that these methods generate clusters neither exceeding nor falling below 10 in comparison to our label number.  

We employ a dynamic addition strategy to determine the size of $\mathcal{D}_{sub}$ for each dataset. Starting with the empty set, we iteratively add 1000 examples at a time, generating the initial label space until no additional labels are generated. Consequently, $\mathcal{D}_{sub}$ contains 3000 documents for AAPD, Reuters-21578, and RCV1-V2; 8000 documents for DBPedia-298; and 14000 documents for Amazon-531.
In the label space improvement phase, by ranking the top entailment scores in ascending order, we select a subset of chunks $S$ with comparatively lower entailment scores. 
To ensure consistency with $\mathcal{D}_{sub}$, we control the size of $S$ proportionally. 
For AAPD, Reuters-21578, and RCV1-V2 datasets, we select the top 500 examples. 
For DBPedia-298, the subset size is set at 1,000, and for Amazon-531, we choose 1,500 examples. 

We consider labels existing in less than 1\% of the dataset as long-tail labels. Therefore, we select all keyphrases that occur in less than 1\% of $\mathcal{P}$. By calculating the semantic similarity score between these selected keyphrases and all labels in $C$, we identify the highest score for each keyphrase and use the median of these scores as our threshold $\gamma$.
Only labels with a semantic similarity score lower than $\gamma$ compared to existing labels are added to the new label space. 
This methodology ensures a refined and relevant augmentation of the label space. Details on the time and computational resources required for model training and prediction are provided in Appendix~\ref{sec:resource}.


\subsection{Label Space Coverage Results}

We present the coverage of the predicted label space in comparison to topic modeling and keyword extraction baselines, as detailed in Table~\ref{table:coverage}. 
Our method consistently outperforms all baseline approaches. 
Specifically, for the AAPD, RCV1-V2, and DBPedia-298 datasets, we achieve over 60\% coverage of the ground-truth label space, showcasing a noteworthy increase of up to 40\% compared to the baseline methods. These results demonstrate that our method excels at predicting labels that align closely with the ground-truth label space.
For example, our model can generate long-tail labels, a task that is particularly challenging for all baselines.
However, our model exhibits comparatively lower performance on the Reuters-21578 and Amazon-531 datasets. 
Regarding Reuters-21578, the lower performance is due to a higher proportion of long-tail labels and the use of abbreviations in ground-truth labels.
For Amazon dataset, the initially generated label space by \our is only one-third of the ground-truth size. Even after adding more labels in the improvement stage, the predicted label space is still less than half of the ground-truth space, leading to lower coverage scores.

We also use \texttt{Meta-Llama-3-8B-Instruct} for model implementation and test on AAPD and RCV1-V2 datasets. The coverage scores were 75.51\% and 60.19\%, respectively, similar to the results obtained using \texttt{llama-2-13b-chat}. 

\subsection{Zero-shot Text Classification Accuracy}

We present the comprehensive zero-shot performance across all methods in Table~\ref{table:accuracy}. The results unequivocally demonstrate that our framework consistently outperforms nearly all baseline models. Notably, the P@3 scores of X-MLClass surpass those of the baseline methods across all datasets. This observation implies that training the zero-shot classifier for both the keyphrases set and the chunk set, followed by merging the results, enhances the multi-label performance. 
Specifically, our chunk-splitting procedure increases the likelihood of finding the less dominant labels for each document, as these labels may become dominant in smaller chunks. 
Similarly, our approach improves the accurate prediction of tail labels by the classifier, contributing to the overall MLTC performance.  

\begin{table}[t]
\small
  \begin{center}
    \vspace{0mm}
\resizebox{\linewidth}{!}{
    \begin{tabular}{lccr} 
    \toprule
      \textbf{Dataset} & \textbf{Initial} & \textbf{Improvement}  & \textbf{$\Delta$}\\
      \midrule
      AAPD & 44.90 & 77.56 & +32.66\%\\
      Reuters-21578 & 24.44 & 37.78 & +13.34\%\\
      RCV1-V2 & 49.51 & 61.17 & +11.66\%\\
      DBPedia-298 & 55.70 & 67.45 & +11.75\%\\
      Amazon-531 & 23.35 & 38.04 & +14.69\%\\
      \bottomrule
    \end{tabular}
    }
    \caption{Label Coverage Score Improvement Results.}
    \label{table:self-training}
    \vspace{-3mm}
    
  \end{center}
\end{table}

\vspace{-2mm}
\subsection{Label Space Coverage Improvement}
\vspace{-1mm}

\begin{figure}[t]
    \centering
    \includegraphics[width=0.85\linewidth,trim={0 0 0 1.10cm},clip]{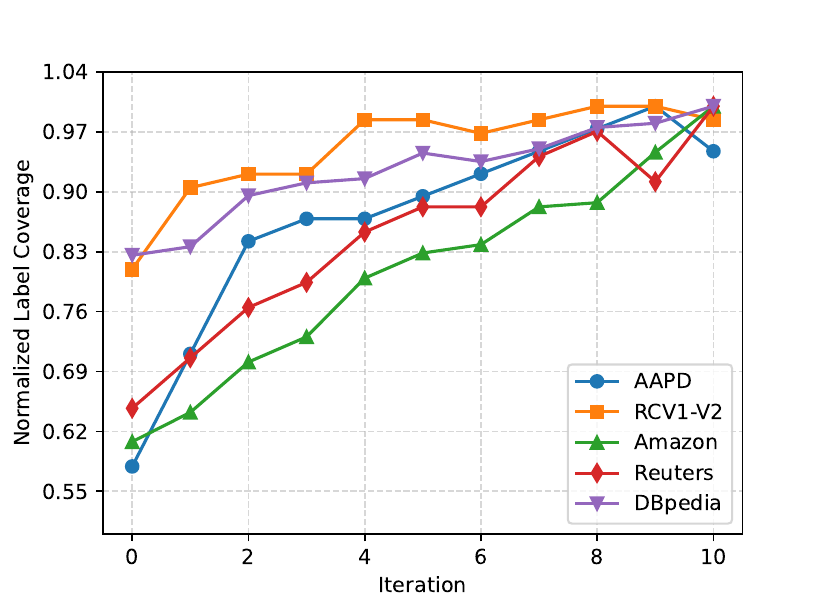}
    \vspace{-3mm}
  \caption{Coverage Improvement across Iterations.}
    \label{fig:selfTraining}
    \vspace{-3mm}
\end{figure}

Table~\ref{table:self-training} shows that iteratively updating the label space leads to an enhancement in label coverage across all datasets. 
Figure~\ref{fig:selfTraining} visually represents the normalized incremental coverage during each iteration across datasets.  
Notably, the improvement is more pronounced for datasets with smaller initial label space sizes or lower initial coverage scores. 
This finding aligns with expectations, as AAPD exhibits significantly smaller label space sizes compared to all other datasets, rendering label space improvement easier. 
For the Reuters-21578 and Amazon-531 datasets, the initial coverage score is low, meaning that many matching labels are not initially included in the label space. This results in a higher potential for improvement by adding these labels.   
Additionally, the criteria for adding new labels must align with all existing ones in the generated label space, presenting a greater challenge in expanding larger label spaces like DBPedia-298.


\subsection{Label Coverage with Human Involvement/Evaluation}
We can seek expert assistance for refinement when generating the initial label space, especially for borderline similar label pairs. Human judgment helps determine whether to keep both labels, typically requiring inspection of only about 30 pairs on average. 
Table~\ref{table:human-invol} shows that the coverage score of initial label space across all datasets is lower without human involvement, as expected.

Our model also encounters challenges in generating labels exactly matching the ground-truth label space. For instance, within Reuters-298, certain ground-truth labels are abbreviations, while our model generates the full-word version, leading to a lower semantic similarity score than the actual score. As shown in Table~\ref{table:human-eval}, the ground-truth label ``acq'' corresponds to our predicted label ``acquisitions,'' possessing identical meanings, yet their semantic similarity score falls below 40\%.
In the Amazon-531 dataset, many ground-truth labels consist of phrases, which complicates the coverage evaluation. Achieving high scores requires precise matches, but predicting similar-meaning phrases with different words is common, resulting in lower scores. 
As evident in Table~\ref{table:human-eval}, ``electrical\_safety'' and ``electronics\_troubleshooting'' are identical labels, but their semantic similarity scores are lower, treated as distinct labels in our setting. Expert evaluation can assist in such cases.

Considering these factors, the actual coverage score of our predicted label space compared to the ground-truth label space is likely higher than the presented result in Table~\ref{table:coverage}.

\subsection{Ablation Study for Amazon-531 Dataset}

The label space for the Amazon-531 dataset is significantly larger than that of the other datasets. To address this discrepancy and enhance label coverage, we increased the number of iterations to add more long-tailed labels. Using the same number of iterations as for the other datasets would result in a final label space only half the size of the ground-truth label space. As depicted in Figure~\ref{fig:amazon}, increasing the number of iterations facilitates the addition of more labels to the predicted label space, leading to an improvement in the coverage score.

\begin{table}[t]
\small
  \begin{center}
\resizebox{\linewidth}{!}{
    \begin{tabular}{lccr} 
    \toprule
      \textbf{Dataset} & \textbf{w/o Human} & \textbf{ w/ Human}  & \textbf{$\Delta$}\\
      \midrule
      AAPD & 40.81 & 44.90 & +4.09\%\\
      Reuters-21578 & 18.89 & 24.44 & +5.55\%\\
      RCV1-V2 & 45.63 & 49.51 & +3.88\%\\
      DBPedia-298 & 46.31 & 55.70 & +9.39\%\\
      Amazon-531 & 21.28 & 23.35 & +2.07\%\\
      \bottomrule
    \end{tabular}
    }
    \caption{Initial Coverage w/wo Human Involvement.}
    \label{table:human-invol}
   \vspace{-3mm}    
  \end{center}
\end{table}

\begin{table}[t]
\small
  \begin{center}
    \scalebox{0.95}{
    \begin{tabular}{cc} 
    \toprule
      \textbf{Ground-truth} & \textbf{Predicted Label}\\
      \midrule
      acq & acquisitions \\
      money-fx & monetary policy \\
      earn & earnings \\
      plug\_play\_video\_games & gaming\_electronics  \\
      electrical\_safety & electronics\_troubleshooting \\
      teether & baby\_dental\_care\\
      \bottomrule
    \end{tabular}
    }
    \caption{Matching pairs between the ground-truth labels and the predicted labels through human evaluation.}
    \label{table:human-eval}
    \vspace{-7mm}
  \end{center}
\end{table}

\section{Conclusion and Future Work}
We attack a novel and challenging problem, open-world MLTC with extremely weak supervision. 
In this scenario, user only provides a brief description for classification objectives without any labels or ground-truth label space. Our LLM-based framework, \our, is designed to overcome this challenge by discovering a practical label space and constructing an MLTC classifier for label prediction. 
Notably, it excels in identifying long-tail labels, arguably the most challenging aspect in MLTC problems.
Our experiment results show that \our surpasses baselines in terms of ground-truth label coverage and exhibits higher zero-shot text classification performance compared to top-ranking models.

Despite our model's success in generating some tail labels, a considerable number of tail labels remain undiscovered. 
Future work should focus on refining our approach to capture more long-tail labels. 
Subsequent studies could explore methodologies tailored for datasets featuring significantly larger label spaces, contributing to the broader applicability of our model.  

\section*{Limitations}
Our work aims to discover the label space from extensive input text documents and then construct a multi-label text classifier. The most formidable challenge in this problem setting revolves around label space construction --- how can we discover the labels, especially the long-tail ones?
Therefore, our primary focus is on developing a novel method to address this challenge; we didn't propose any new zero-shot multi-label text classifier, since it is beyond the scope of this paper.
Given that our proposed \our starts with a subset of documents, its efficacy may be limited for extremely long-tail labels (e.g., those occurring less frequently than 0.0001\% of the documents).  Alternatively, a considerably large subset would be required, potentially incurring significant computational costs from LLM.
While our evaluation includes a diverse set of datasets, there is potential for further extension to more challenging datasets with an exceptionally large label space (e.g., over 1000 different labels are expected). 

\section*{Ethical Considerations}

This paper uses open-source datasets and models for training and evaluation, which is reproducible. We will also release the code upon acceptance. It is important to note that the keyphrases generated by LLMs may vary with each run, potentially leading to minor differences in results compared to those presented. Additionally, our evaluation toolkit uses OpenAI models, which may impact reproducibility. 

\bibliography{custom.bib}

\appendix

\begin{figure}[ht]
    \centering
    \includegraphics[width=0.8\linewidth]{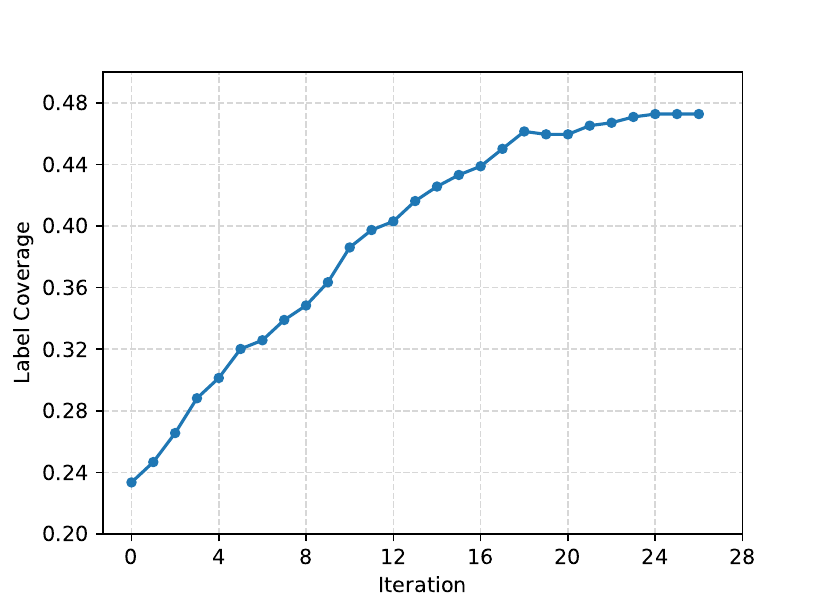}
  \caption{Improvement of Label Coverage for Amazon-531 by increasing the number of iterations.}
    \label{fig:amazon}
\end{figure}

\section{Prompt Templates for Dominant Label}
\label{sec:prompt}
Code~\ref{dom_ppt} is the prompt we use to find the dominant label for the selected document.

\section{Prompt Templates for Generating Keyphrases}
\label{sec:phrases}
Code~\ref{keyphrase_ppt} provides an example of the prompt used to generate keyphrases for a selected chunk of the Amazon-531 dataset. Users can help us define the objective with examples. For example, the coarse-grained objectives look like ``games'' and ``animals'', while the corresponding fine-grained objectives are ``trading\_card\_games'' and ``reptiles''.

\section{Label Space refinement with human involvement}
\label{sec:human-involve}

Human experts play a crucial role in refining the label space generated by LLM. For instance, when the cosine similarity score between two labels falls between 0.50 and 0.75, indicating a certain degree of semantic similarity, human intervention is preferred to determine whether these labels are synonymous. Synonyms need to be identified and treated accordingly, with one of them being removed from the label space. To specify, we show them with an instruction ``\textit{Do label pairs have similar meanings in the text classification problem? If Yes, please output the label that we should delete}.'' However, there is also the case that these two labels may represent concepts from different scopes; for example, ``health\_care'' and ``health\_personal\_care.'' In such instances, human judgment is necessary to detect and treat them as separate labels.

Furthermore, some predicted labels may contain multiple meanings, necessitating human intervention to split them into distinct labels. For instance, if a predicted label is ``computer vision and machine learning,'' it is evident that the label should be divided into two separate labels. These judgments require human expertise for accurate and context-aware decisions.

\section{Datasets Detailed Information}
\label{sec:dataset}

\begin{itemize}[nosep,leftmargin=*]
    \item \textbf{{AAPD}}~\cite{Yang2018SGMSG} contains computer science papers. The labels are research topics.
    \item \textbf{Reuters}-21578~\cite{Sebastiani2003AnAO} is a collection of news articles from the Reuters financial newswire service in 1987. The labels are the news topics.
    \item \textbf{RCV1-V2}~\cite{lewis2004rcv1} contains categorized newswire articles by Reuters Ltd. The labels are the news topics.
    \item \textbf{DBPedia}-298~\cite{lehmann2015dbpedia} are extracted from Wikipedia articles. The labels are the article categories. 
    \item \textbf{Amazon}-531~\cite{McAuley2013HiddenFA} encompasses product reviews and associated metadata. The labels are the product tags. 
\end{itemize}

\section{GPT Instructions for Verifying Matching Pairs}
\label{sec:gpt}
Code~\ref{pair_ppt} is the instruction we used to verify matching pairs. 

\section{Time and Computational Resources Required for Model Training and Prediction}
\label{sec:resource}
The number of training examples we select varies based on the scope of each dataset. On average, we prompt LLM on approximately 18,000 chunks for each dataset. We use a single 80GB A100 for LLM prompting, and the process consumes less than 35GB of GPU memory. Each prompting only takes less than 1 second to execute. Thus, on average, each dataset requires less than 4 hours for LLM prompting, making it a manageable cost.

During the text classification and label space improvement step, the majority of the computation cost arises from calculating the textual entailment score for each chunk. However, the entailment model is lightweight, occupying only around 3GB of GPU memory. When dealing with large label spaces, nearest-neighbor techniques can be applied to trim down label candidates, reducing the time required for calculating entailment scores. On average, we require approximately 2 hours to complete one iteration. However, users have the flexibility to choose the number of iterations needed to update the label space. Given that our model already outperforms the baseline with the initial label space, only a few iterations are necessary, making both time and cost manageable.

\begin{figure*}[t]
\begin{minipage}{1.0\textwidth}
\begin{lstlisting}[language=Python, label=dom_ppt, caption={Prompt to find the dominant label for the selected document}]
sys_prompt = "You are a poetic assistant, skilled in explaining complex programming concepts with creative flair."
user_prompt = "
    Which label in the label space {true_label_array} is the dominant label that covers more than 50% of the below content?     
    {documents}
    
    Please output the dominant label only if exist or output \'NO\' if there are no dominant labels. 
    "
prompt = "{sys_prompt} \n {user_prompt}"

\end{lstlisting}
\end{minipage}
\end{figure*}

\begin{figure*}[t]
\begin{minipage}{1.0\textwidth}
\begin{lstlisting}[language=Python, label=keyphrase_ppt, caption={Prompt templates to generate keyphrases}]
#Amazon-531 keyphrases generation template
sys_prompt = "<s>[INST] <<SYS>>
    You are a helpful, respectful and honest assistant for labeling topics.
    <</SYS>>."
example_prompt = "
    The following is a customer review of a product bought from Amazon.     
    [documents]
    Based on the topic information mentioned above, the coarse-grained keyphrases are formatted as {user classification objective}, while the fine-grained keyphrases are formatted as {user classification objective}.
    "
main_prompt = "
    [INST]   
    Based on the information about the topic above, please find two coarse-grained and two fine-grained keyphrases for the example. 
    [DOCUMENTS] 
    Please only return the keyphrases in one line using the format below:
    [/INST] [keyphrase] and [/keyphrase].
    "
prompt = "{sys_prompt} \n {example_prompt} \n {main_prompt}"

\end{lstlisting}
\end{minipage}
\end{figure*}

\begin{figure*}[t]
\begin{minipage}{1.0\textwidth}
\begin{lstlisting}[language=Python, label=pair_ppt, caption={Prompt templates for verifying match pairs}]
sys_prompt = "You are an expert in text classification, with specialized skills in discerning matching pairs for labels."
user_prompt = "
    Given that we have established matching pairs such as 
    "\'Machine learning\' and \'artificial intelligence\'", 
    "\'Computational Geometry\' and \'Algebraic Geometry\'", 
    "\'Physics and Society\' and \'Physics\'",       //optional
    "\'teether\' and \'baby_dental_care\'",          //optional
    when using util.dot_score to measure semantic similarity between tokens, would you consider {ground_truth} and {prediction} as a matching pair in a text classification problem? 
    
    Please respond with Yes or No. 
    "
prompt = "{sys_prompt} \n {user_prompt}"

\end{lstlisting}
\end{minipage}
\end{figure*}

\end{document}